\documentclass[letterpaper, 10 pt, conference]{ieeeconf}  % Comment this line out if you need a4paper

\IEEEoverridecommandlockouts                              % This command is only needed if 
\usepackage{amsmath,amsfonts}

\usepackage{array}

\usepackage[caption=false,font=normalsize,labelfont=sf,textfont=sf]{subfig}

\usepackage{textcomp}
\usepackage{stfloats}
\usepackage{url}
\usepackage{verbatim}
\usepackage{graphicx}

\usepackage[ruled,linesnumbered]{algorithm2e}
\usepackage{algpseudocode}
\usepackage{algorithmicx}
\usepackage{algcompatible}

\usepackage{amssymb}

\usepackage{subcaption} 
\usepackage{adjustbox}

\usepackage{booktabs}
\usepackage{multirow}

\usepackage{xcolor}

\usepackage{cite} 
\usepackage[colorlinks=true, linkcolor=blue, citecolor=blue, urlcolor=blue]{hyperref}

\title{\LARGE \bf
D²-LIO: Enhanced Optimization for LiDAR-IMU Odometry Considering Directional Degeneracy
}

\author{Guodong Yao, Hao Wang, and Qing Chang% <-this % stops a space
\thanks{This work was supported by the National Key R\&D Program of China (No. 2024YFC3015605).}% <-this % stops a space
\thanks{Guodong Yao, Hao Wang, and Qing Chang (Corresponding Author) are with the School of Electronic and Information Engineering, Beihang University, Beijing 100191, China. 
        E-mail: guodongyao47@gmail.com; whbuaa@qq.com; buaahao@126.com.}%
}

\begin{document}

\maketitle
\thispagestyle{empty}
\pagestyle{empty}

%%%%%%%%%%%%%%%%%%%%%%%%%%%%%%%%%%%%%%%%%%%%%%%%%%%%%%%%%%%%%%%%%%%%%%%%%%%%%%%%
\begin{abstract}

LiDAR-inertial odometry (LIO) plays a vital role in achieving accurate localization and mapping, especially in complex environments. However, the presence of LiDAR feature degeneracy poses a major challenge to reliable state estimation. To overcome this issue, we propose an enhanced LIO framework that integrates adaptive outlier-tolerant correspondence with a scan-to-submap registration strategy. The core contribution lies in an adaptive outlier removal threshold, which dynamically adjusts based on point-to-sensor distance and the platform’s motion amplitude. This mechanism improves the robustness of feature matching in varying conditions. Moreover, we introduce a flexible scan-to-submap registration method that leverages IMU data to refine pose estimation, particularly in degenerate geometric configurations. To further enhance localization accuracy, we design a novel weighting matrix that fuses IMU preintegration covariance with a degeneration metric derived from the scan-to-submap process. Extensive experiments conducted in both indoor and outdoor environments—characterized by sparse or degenerate features—demonstrate that our method consistently outperforms state-of-the-art approaches in terms of both robustness and accuracy.

\end{abstract}

%%%%%%%%%%%%%%%%%%%%%%%%%%%%%%%%%%%%%%%%%%%%%%%%%%%%%%%%%%%%%%%%%%%%%%%%%%%%%%%%
\section{INTRODUCTION}
\label{sec:introduction}

Recent advancements in robotics have substantially enhanced autonomous navigation and mapping capabilities, primarily driven by innovations in sensing technologies and algorithmic development~\cite{papachristos2019autonomous}. Among these, simultaneous localization and mapping (SLAM) has emerged as a cornerstone technology, enabling real-time localization and map-building in unknown environments~\cite{zhang2014loam, shan2020lio,xu2022fast,yuan2022voxelmap,jiang2022pvlio}. SLAM plays a crucial role across a wide spectrum of applications, including autonomous vehicles, aerial drones, and planetary rovers operating in global positioning system (GPS)-denied or complex terrains~\cite{ebadi2023present}.

Among various sensor modalities,  light detection and ranging (LiDAR) has gained prominence for its ability to capture high-precision 3D spatial information. LiDAR-based SLAM systems excel in generating detailed maps, enabling robust navigation in both structured and expansive outdoor environments~\cite{zhang2014loam,shan2020lio,xu2022fast,yuan2022voxelmap,jiang2022pvlio}. However, despite their advantages, LiDAR-based systems face significant challenges in environments that lack distinctive geometric features—such as tunnels, forests, and open fields—where SLAM performance may degrade~\cite{ebadi2023present}.

A key limitation in these scenarios is degeneration, a phenomenon that occurs when point cloud data lacks sufficient geometric variation to constrain full 6-DoF pose estimation. This often arises in feature-scarce or repetitive environments where methods like iterative closest point (ICP) struggle to align consecutive scans~\cite{ebadi2021dare,hatleskog2024probabilistic,liao2025real,tuna2024informed,tuna2023x,lee2024genz,besl1992method}. As a result, optimization becomes ill-conditioned, leading to pose drift and accumulated mapping errors. Detecting and mitigating degeneracy is essential to maintaining SLAM accuracy under such conditions.

To address these challenges, recent research has explored several strategies. Some approaches dynamically adjust the contribution of point cloud features based on their reliability, improving alignment in degenerate areas~\cite{tuna2023x}. Others use probabilistic techniques to assess uncertainty in pose estimation and prevent overconfident updates in weakly constrained directions~\cite{hatleskog2024probabilistic}. Geometric analyses, such as examining the eigenvalues of the Hessian matrix, have also been applied to identify and respond to degenerate conditions. Additionally, denoising methods and sensor fusion—incorporating cameras or inertial measurement units (IMUs)—have been employed to provide complementary constraints and enhance robustness~\cite{liao2025real}.

Despite these advances, existing methods often rely on empirically tuned parameters or augmentations to traditional ICP frameworks, which remain sensitive to noise and prone to local minima. And the root causes of degeneration—such as geometric ambiguity and sensor noise—remain underexplored, limiting the generalizability of current solutions. 
Furthermore, while integrating additional sensing modalities can alleviate the effects of LiDAR degeneracy, the fundamental challenge lies in developing more robust and reliable strategies to directly address LiDAR degeneracy, rather than bypassing it unless absolutely necessary. There is a pressing need for a reimagined registration framework that is both noise-resilient and capable of adapting to geometric degeneracy.

\begin{figure*}[htp]
  \centering
  \includegraphics[width=0.95\textwidth]{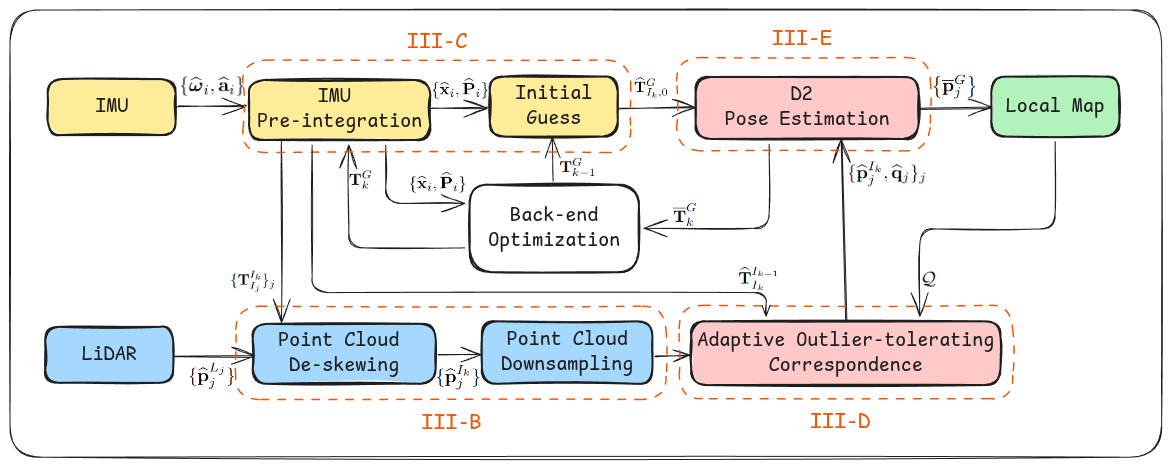}
  \captionsetup{justification=raggedright, singlelinecheck=false}
  \caption{System overview of D²-LIO}
  \label{fig:D2-LIO}
\end{figure*}

\textbf{This paper presents a comprehensive analysis of LiDAR SLAM degeneration in unstructured environments and proposes a novel LiDAR-IMU front-end matching pipeline to enhance robustness under such conditions,} as shown in Fig. \ref{fig:D2-LIO}.

We examine the degeneration problem from three perspectives: \begin{itemize} 
\item \textit{Low observability of state variables}: As illustrated in Fig. 3 of ~\cite{gelfand2003geometrically}, certain environment types lead to unobservable variables that degrade SLAM performance. 
\item \textit{Point cloud mismatches}: Unstructured environments tend to cause point cloud mismatches during the ICP process, resulting in errors in the formulation of the optimization 
\item \textit{Ill-conditioned optimization}: In these scenarios, the ICP cost function tends to exhibit flat minima, making the optimization process highly sensitive and prone to error.\end{itemize}

The low observability of state variables is fundamentally dictated by environmental geometry and thus lies beyond algorithmic control. To tackle the remaining two challenges, we propose a dynamic, per-point thresholding mechanism for robust correspondence filtering, and an adaptive fusion framework that integrates geometric and inertial constraints to strengthen registration under degenerate conditions.

\noindent{\bf Our contributions are as follows:}
\begin{enumerate}
\item{
    We introduce a dynamic, per-point distance thresholding mechanism that selectively filters out invalid correspondences by accounting for both the sensor-to-point distance and platform motion, effectively improving matching robustness in degenerate scenarios.
}
\item{We develop an optimization framework that fuses geometric and inertial constraints using an adaptive weighting scheme, enabling robust pose estimation even under weak geometric conditions.}
\item{
    We develop an improved LIO front-end and integrate it into a modified LIO-SAM framework for system-level evaluation. By benchmarking the proposed method against existing LiDAR-SLAM systems, we demonstrate its enhanced robustness in feature-scarce environments.
}
\end{enumerate}

\section{RELATED WORK}

\subsection{Robust Point Cloud Registration}
Point cloud registration is a fundamental component of LiDAR-based SLAM, with the ICP algorithm being the most widely used approach~\cite{besl1992method}. ICP iteratively estimates the transformation between point clouds by establishing correspondences—typically via nearest-neighbor searches—and minimizing a cost function that reflects alignment error. To better handle the sparsity and structure of point cloud data, various error metrics such as point-to-point~\cite{besl1992method} and point-to-plane~\cite{censi2008icp} have been employed to improve registration stability and accuracy. However, traditional ICP is sensitive to outliers and noise, which can severely degrade performance. To address this, robust M-estimators—such as Cauchy, and Huber—are incorporated via loss functions to down-weight outlier influence and enhance robustness~\cite{vizzo2023kiss}. To further tackle ICP’s tendency to get trapped in local minima, globally optimal methods such as TEASER~\cite{yang2020teaser}, based on graduated non-convexity, have gained traction for their ability to handle high outlier ratios and ensure reliable registration.

Beyond classical ICP-based methods, alternative registration strategies have been developed to improve robustness in challenging environments. Redundancy-aware approaches such as RMS~\cite{petracek2024rms} reduce point redundancy to enhance real-time pose estimation under geometric degeneracy. Other directions include feature-based methods that extract salient structures~\cite{behley2018efficient}, probabilistic models grounded in statistical inference~\cite{myronenko2010point}, and learning-based techniques that improve adaptability across diverse scenarios~\cite{bai2021pointdsc}.

\subsection{LiDAR-inertial Odometry}
LiDAR-inertial odometry (LIO) is central to SLAM and is typically implemented through either optimization-based or filtering-based frameworks. Optimization-based methods—exemplified by LOAM~\cite{zhang2014loam} and its successors such as LIO-SAM~\cite{shan2020lio} employ scan matching strategies to align sparse geometric features and estimate poses. These approaches are widely adopted for their balance of real-time performance and accuracy.

Filtering-based approaches, such as Fast-LIO~\cite{xu2022fast}, utilize aniterated extended state Kalman filter (IESKF) to tightly couple LiDAR and IMU data. Unlike feature-based methods, Fast-LIO eliminates the need for explicit feature extraction, yielding high precision and efficiency, particularly in dynamic or resource-constrained environments.

In parallel, semantic SLAM has emerged to enhance scene understanding and robustness by integrating object-level or semantic features~\cite{chen2019suma++}. While promising for handling dynamic scenes and improving obstacle recognition, these approaches are often limited by their computational demands, which hinder their viability for real-time applications.

\subsection{Degeneracy Detection and Mitigation}

Degeneracy in LiDAR SLAM occurs when the optimization becomes ill-conditioned, typically in environments with repetitive or sparse geometric features, such as corridors or open fields. This leads to ambiguous pose estimation and unreliable mapping results. To address this, a variety of detection and mitigation strategies have been proposed.

Degeneracy detection methods generally fall into three categories: geometric, optimization-based, and data-driven approaches~\cite{tuna2023x}. Geometric methods analyze the interaction between point cloud structure and registration cost function. For example, X-ICP incorporates a geometry-aware detection scheme that evaluates degeneracy in both rotational and translational subspaces without requiring prior maps~\cite{tuna2023x}. Optimization-based methods assess metrics such as the condition number of the Hessian matrix~\cite{ebadi2021dare} to identify ill-conditioned states. While insightful, these approaches often rely on heuristic thresholds and are sensitive to scene variability. Data-driven techniques, including support vector machines~\cite{nobili2018predicting} and entropy-based metrics~\cite{gao2020prediction}, predict degeneracy directly from LiDAR data but often require extensive training and incur high computational costs.

Mitigation strategies are typically categorized as passive or active~\cite{tuna2024informed}. Passive methods introduce complementary sensors like IMUs or visual odometry to provide external constraints when degeneracy is detected~\cite{shan2020lio},~\cite{khattak2020complementary}. Active approaches modify the optimization pipeline itself—for instance, solution remapping redirects the solver toward well-conditioned directions~\cite{zhang2016degeneracy}, while regularization terms or hard constraints are introduced to stabilize optimization under degenerate conditions~\cite{tuna2024informed}. X-ICP~\cite{tuna2023x}, for example, applies directional constraints to reduce drift. However, these solutions may still fail in extreme cases where degeneration is too severe to recover from, especially when divergence occurs early in the optimization process.

\section{METHODOLOGY}

\subsection{Notations and Preliminaries}
We begin by introducing the frames of reference and notation employed throughout this paper. The global frame is denoted as $\mathcal{G}$, and the robot body frame is denoted as $\mathcal{B}$. For simplicity, we assume that the IMU frame coincides with the robot body frame. The robot state $\mathbf{x}$ is expressed as:
\begin{equation}
  \mathbf{x} = \begin{bmatrix} 
    \mathbf{R}^{\top}, \mathbf{t}^{\top}, \mathbf{v}^{\top}, \mathbf{b}_g^{\top}, \mathbf{b}_a^{\top}, \mathbf{g}^{\top} 
  \end{bmatrix}^{\top},
\end{equation}
where $\mathbf{R} \in SO(3)$ is the rotation matrix, $\mathbf{t} \in \mathbb{R}^3$ is the position vector, $\mathbf{v} \in \mathbb{R}^3$ is the velocity vector, $\mathbf{b}_g \in \mathbb{R}^3$ and $\mathbf{b}_a \in \mathbb{R}^3$ are the gyroscope and accelerometer biases, respectively, and $\mathbf{g} \in \mathbb{R}^3$ is the gravity vector.

The transformation matrix $\mathbf{T} \in SE(3)$ of body state at time \( \mathrm{t}_k \) is represented as: $\mathbf{T}_{I_k}^G = [\mathbf{R}_{I_k}^G, \mathbf{t}_{I_k}^G]$, which maps the body frame $\mathcal{B}$ to the global frame $\mathcal{G}$. And extrinsic transformation from LiDAR to IMU is $\mathbf{T}_L^I$. LiDAR point cloud at time $\mathrm{t}_k$ is $\widehat{\mathcal{P}}_k$, where the $j$-th point is $\widehat{\mathbf{p}}^{L_j}_j$.

\subsection{Preprocessing of LiDAR Point Cloud}
To enable accurate and efficient point cloud registration, raw LiDAR data are first preprocessed through two key steps: de-skewing and down-sampling.

\subsubsection{Point Cloud De-skewing}
Sequential LiDAR scanning causes points within a scan to have varying timestamps, resulting in motion distortion. We correct this by transforming each point to a common IMU frame at time \( t_k \):
\begin{equation}
    \widehat{\mathbf{p}}_{j}^{I_k}=\mathbf{T}_{I_j}^{I_k}\mathbf{T}_{L}^{I} \widehat{\mathbf{p}}^{L_j}_j.
\end{equation}

\subsubsection{Point Cloud Down-sampling}
To enhance real-time performance, we adopt voxel grid filtering—a simple yet effective method—for point cloud down-sampling. While it may inevitably sacrifice some fine-grained geometric details, this choice reflects a well-considered and pragmatic trade-off between computational efficiency and spatial fidelity, consistent with the real-time constraints of our system.

\subsection{Discrete Model of IMU Preintegration}
The angular velocity and acceleration measurements obtained from an IMU at time $t$ are given by the following expressions:
\begin{equation}
  \begin{aligned}
  \widehat{\boldsymbol{\omega}}_{t} & =\boldsymbol{\omega}_{t}+\mathbf{b}_{g,t}+\mathbf{n}_{g,t} \\
  \widehat{\mathbf{a}}_{t} & =\mathbf{R}_{t}^{\top}\left(\mathbf{a}_{t}-\mathbf{g}_{t}^{G}\right)+\mathbf{b}_{a,t}+\mathbf{n}_{a,t}
  \end{aligned},
\end{equation}
where $\boldsymbol{\omega}_{t}$ and $\mathbf{a}_{t}$ represent real angular velocity and acceleration, which are influenced by a slowly varying bias $\mathbf{b}_{t}$ and white noise $\mathbf{n}_{t}$.

Based on the $\boxplus$ operation defined in \cite{xu2022fast}, the discrete model of state estimation can be formulated by:
\begin{equation}
  \mathbf{x}_{i+1} = \mathbf{x}_{i} \boxplus \left( \Delta t \mathbf{f}(\mathbf{x}_{i}, \mathbf{u}_{i}, \mathbf{w}_{i}) \right),
\end{equation}
where $i$ denotes the index of IMU measurements. The system function $\mathbf{f}$, input vector $\mathbf{u}$, and noise vector $\mathbf{w}$ follow a formulation similar to that in the Fast-LIO framework \cite{xu2022fast}, where $\mathbf{f}$ models the nonlinear motion dynamics, $\mathbf{u}$ denotes the raw IMU measurements, and $\mathbf{w}$ represents the corresponding measurement noise.

The propagated state $\hat{\mathbf{x}}_{i}$ and its covariance $\widehat{\mathbf{P}}_{i+1}$ are computed iteratively through the following equations:
\begin{equation}
    \begin{aligned}
        \widehat{\mathbf{x}}_{i+1} &= \widehat{\mathbf{x}}_{i} \boxplus\left(\Delta t \mathbf{f}\left(\widehat{\mathbf{x}}_{i}, \mathbf{u}_{i}, \mathbf{0}\right)\right) ; \quad \widehat{\mathbf{x}}_{0}=\overline{\mathbf{x}}_{k-1} \\
        \widehat{\mathbf{P}}_{i+1} &= \mathbf{F}_{\mathbf{x}} \widehat{\mathbf{P}}_{i} \mathbf{F}_{\mathbf{X}}^{\top} + \mathbf{F}_{w} \mathbf{Q} \mathbf{F}_{w}^{\top} ; \quad \widehat{\mathbf{P}}_0=\overline{\mathbf{P}}_{k-1}
    \end{aligned},
\end{equation}
where $\mathbf{Q}$ is the covariance of the white noise vector $\mathbf{w}$, and $\mathbf{F}_{\mathbf{x}}$ and $\mathbf{F}_{\mathbf{w}}$ are Jacobian matrices. 

\subsection{Adaptive Outlier-tolerating Correspondence}

Outlier rejection is critical to ensuring the accuracy and robustness of point cloud registration. A common approach is to apply a fixed threshold, rejecting correspondences whose spatial distances exceed a pre-defined limit. However, this strategy has inherent limitations: as the distance from a point to the LiDAR increases, the spatial deviation due to motion also grows, leading to a higher rejection rate for distant points compared to nearby ones. This distance-dependent bias can result in the loss of valuable structural information, particularly in sparse or large-scale environments.

To address this issue, we propose an adaptive thresholding strategy inspired by the KISS-ICP framework~\cite{vizzo2023kiss}. While KISS-ICP computes a global threshold based solely on ego-motion for each scan registration, our method extends this by introducing per-point thresholds that also incorporate the distance of each point to the sensor. This refinement ensures that rejection probability remains consistent across spatial scales, improving robustness against motion-induced discrepancies.

Specifically, for each input point $\widehat{\mathbf{p}}_j^{I_k}$, we identify its closest correspondence from the local submap $\mathcal{Q}$:
\begin{equation}
  \widehat{\mathbf{q}}_{j} = \arg \min_{\mathbf{q} \in \mathcal{Q}} \left\| \widehat{\mathbf{T}}_{I_k}^{G} \widehat{\mathbf{p}}_j^{I_k} - \mathbf{q} \right\|_2.
\end{equation}
The correspondence is accepted only if the distance is within a point-specific threshold $\epsilon_j$, defined as:
\begin{equation}
    \epsilon_{j} = \left\| \widehat{\mathbf{t}}_{I_k}^{I_{k-1}} \right\|_2 + 2 \left\| \widehat{\mathbf{p}}_j^{I_k} \right\|_2 \sin \left( \frac{1}{2} \left\| \text{Log} \left( \widehat{\mathbf{R}}_{I_k}^{I_{k-1}} \right)^\vee \right\| \right).
\end{equation}
Here, the first term represents the translational displacement, while the second accounts for rotational displacement scaled by the distance of the point to the sensor. This formulation captures the fact that rotation-induced deviations increase with range, allowing the rejection threshold to adapt accordingly.

Despite its per-point computation, the method remains lightweight and efficient, while efficiently improving tolerance to motion-induced outliers across varying ranges.

\subsection{Pose Estimation Considering Directional Degeneracy}

\subsubsection{ICP Algorithm and Degeneracy Analysis}
The front-end matching process is formulated as an optimization problem, aiming to estimate the optimal transformation $\mathbf{T}^\star$ that minimizes the following cost function:
\begin{equation}
  \label{eq:original_optimization_function}
  c(\mathbf{T}) = \sum_{j=1}^{N} \rho\left(\left\|\mathbf{e}_{j}(\mathbf{T})\right\|^2\right),
\end{equation}
where $\mathbf{e}_{j}(\mathbf{T}) = \mathbf{T} \widehat{\mathbf{p}}_{j}^{I_k} - \widehat{\mathbf{q}}_j$ denotes the matching error between the transformed input point $\widehat{\mathbf{p}}_{j}^{I_k}$ and the corresponding reference point $\widehat{\mathbf{q}}_j$, regularized by a robust kernel function $\rho(\cdot)$. Following the iterative procedure of the ICP algorithm~\cite{besl1992method}, the objective function is optimized to a local minimum.

In geometrically degenerate scenarios, however, the cost function frequently exhibits ill-conditioning (see Section~\ref{sec:introduction}), characterized by flat or ambiguous minima that compromise convergence reliability.

Degeneracy analysis can be performed by linearizing the original state estimation problem in (\ref{eq:original_optimization_function}), and reformulating it as the following least-squares problem~\cite{zhang2016degeneracy}:
\begin{equation}
  \label{eq:least_squares}
  \arg \min _{\mathbf{x}}\|\mathbf{A x}-\mathbf{b}\|^2.
\end{equation}
Through the eigen decomposition of $\mathbf{A}^\top \mathbf{A}$, the eigenvalues $\lambda_i$ serve as quantitative indicators of degeneracy severity along their associated eigenvectors $\mathbf{v}_i$, whereby diminishing magnitudes of $\lambda_i$ signify increasingly ill-constrained directions, thereby exacerbating the system’s vulnerability to noise and perturbations in those subspaces, and increasing the risk of convergence toward erroneous solutions.

\subsubsection{Degeneracy-Aware Regularization Strategy}

To mitigate the aforementioned challenges, we enforce reinforced constraints along directions exhibiting pronounced degeneracy, thereby curbing the iterative solution’s tendency to diverge substantially from the true state throughout the optimization process. Concretely, this is achieved by augmenting the original objective function with an additional regularization term formulated as follows:
\begin{equation}
  \frac{1}{\lambda_{i}}\left\|\left(\mathbf{x}-\hat{\mathbf{x}}_{k}\right) \cdot \mathbf{v}_{i}\right\|^2,
\end{equation}
which ensures that the optimal solution is “not far away” from the initial guess $\hat{\mathbf{x}}_{k}$ along the direction $\mathrm{v}_{i}$. The regularization strength varies inversely with $\lambda_{i}$, thereby enforcing stronger constraints in highly degenerate directions (i.e., smaller $\lambda_{i}$), while naturally relaxing the penalty as the local observability improves (i.e., larger $\lambda_{i}$). Summing over all degenerate directions yields the total regularization term:
\begin{equation}
  \sum_{i=1}^n \frac{1}{\lambda_i}\left\|\left(\mathbf{x}-\hat{\mathbf{x}}_{k}\right) \cdot \mathbf{v}_{i}\right\|^2,
\end{equation}
which can be compactly rewritten as a norm weighted by the inverse of the matrix $\mathbf{A}^{\top} \mathbf{A}$:
\begin{equation}
  \label{original_resiudual_sum}
  \left\|\mathbf{x}-\widehat{\mathbf{x}}_{k}\right\|_{\left(\mathbf{A}^{\top} \mathbf{A}\right)^{-1}}^2.
\end{equation}

Considering the covariance matrix of the IMU pre-integration, which reflects the confidence level of the predicted state $\hat{\mathbf{x}}_{k}$, the final weight matrix $\mathrm{W}$ in~(\ref{original_resiudual_sum}) should be formulated by jointly accounting for both the IMU uncertainty and the degree of degeneracy in the scan-to-submap registration:
\begin{equation}
  \label{eq:weight_final}
  \mathbf{W}=\left(\mathbf{A}^{\top} \mathbf{A} \widehat{\mathbf{P}}_{\mathrm{{\mathbf{T}}},k}\right)^{-1},
\end{equation}
where $\widehat{\mathbf{P}}_{\mathrm{T},k}$ denotes the covariance matrix associated with the pose estimate obtained via IMU preintegration. This fusion strategy ensures a balanced integration of the IMU-derived prior and the geometric constraints from the point cloud, resulting in a pose estimation that accounts for the uncertainties inherent in both sources of information.

A noteworthy technical consideration lies in the inherent scale disparity between the rotational and translational components of the Hessian matrix~\cite{tuna2023x}, which necessitates a separate eigenspace analysis for each. Accordingly, the final objective function for LiDAR scan matching is formulated as follows:
\begin{equation}
  \label{eq:optimization_function}
  \arg \min _{\mathbf{T}} 
  \sum_{j=1}^{\mathrm{N}} \rho\left(\left\|\mathbf{e}_{j}(\mathbf{T})\right\|^2\right)
  +
  \mathrm{w}  ( \left\|\mathbf{e}_{r}(\mathbf{T})\right\|_{\mathbf{W}_r}^2 
  + 
  \left\|\mathbf{e}_{t}(\mathbf{T})\right\|_{\mathbf{W}_r}^2)
\end{equation}
where, $\mathbf{e}_{r}(\mathbf{T})$ and $\mathbf{e}_{t}(\mathbf{T})$ denote the rotational and translational residuals, respectively, characterizing the deviation between the current pose $\mathbf{T}$ and the initial guess of pose $\hat{\mathbf{T}}_k$, and 
\begin{equation}
  \label{eq:seperated weight}
  \begin{gathered}
      \mathbf{W}_r = \left(\mathbf{A_r}^{\top} \mathbf{A_r} \widehat{\mathbf{P}}_{r,k}\right)^{-1}
      \\
      \mathbf{W}_t = \left(\mathbf{A_t}^{\top} \mathbf{A_t} \widehat{\mathbf{P}}_{t,k}\right)^{-1},
  \end{gathered}
\end{equation}
which account for the rotational and translational uncertainties embedded in the Hessian matrix, as well as the prior pose uncertainties derived from the initial estimation.

The complete procedure of pose estimation is outlined in Algorithm 1.

\begin{algorithm}
    \caption{Pose Estimation}
    \label{Algorithm:Pose Estimation}
    \SetAlgoLined
    \SetKwInOut{Input}{input}
    \SetKwInOut{Output}{output}
    \DontPrintSemicolon
    \Input{Last output $\overline{\mathbf{T}}^{G}_{k-1}$,\linebreak
           LiDAR raw points in current scan, \linebreak
           IMU input $\{\widehat{\boldsymbol{\omega}}_i, \widehat{\mathbf{a}}_i\}$ during current scan.}
    
    \Output{Current optimal estimate $\overline{\mathbf{T}}^{G}_{k}$,\linebreak
            The transformed LiDAR points $\left\{ \overline{\mathbf{p}}^{G}_{j}\right\}.$}
    
    Forward propogation to obtain state prediction $\widehat{\mathbf{T}}_{I_k, 0}^{G}$
    
    $\kappa=-1, \widehat{\mathbf{T}}_{k}^{\kappa=0}=\widehat{\mathbf{T}}_{I_k, 0}^{G}$
    
    \For  {$\kappa < iterNumMax$} {

        Compute ${\mathbf{A}}_{k}^\kappa$ via \eqref{eq:least_squares}

        Compute ${\mathbf{W}}_{k}^\kappa$ via \eqref{eq:seperated weight}
        
        Compute the pose estimation $\widehat{\mathbf{T}}_{k}^{\kappa+1}$ via \eqref{eq:optimization_function}
        
        \If { $\left\|\widehat{\mathbf{T}}_{k}^{\kappa+1} \boxminus \widehat{\mathbf{T}}_{k}^\kappa\right\| < \epsilon $ }   {
            return    
        }
    }   
    $\overline{\mathbf{T}}^{G}_{k}=\widehat{\mathbf{T}}_{k}^{\kappa+1}$
    
    Obtain the transformed LiDAR points $\left\{\overline{\mathbf{p}}^{G}_{j}\right\}$
\end{algorithm}

\section{EXPERIMENTS}

\subsection{Datasets and Experimental Setup}

We evaluate our method on three publicly available datasets: GEODE~\cite{chen2024heterogeneous}, SubT-MRS~\cite{zhao2024subt}, and ECMD~\cite{chen2023ecmd}. These datasets encompass diverse and challenging environments, including indoor and outdoor scenes with varying degrees of geometric degradation, sensor modalities, and platform types. GEODE focuses on degraded geometries across diverse LiDAR configurations, SubT-MRS provides long-term, multimodal data from extreme real-world conditions, and ECMD targets autonomous driving with degenerate environments like highways and bridges featuring repetitive structures and sparse geometry. Together, they form a comprehensive benchmark to assess SLAM robustness under complex real-world conditions.

All comparative methods were carefully calibrated for LiDAR-IMU extrinsics and IMU intrinsics according to each dataset's recommended settings, while other parameters were kept at their default values. To ensure a fair comparison, all experiments were conducted on the same platform under consistent conditions. Furthermore, performance was evaluated using the absolute trajectory error (ATE) metric, which was computed with the open-source tool evo~\cite{grupp2017evo}.

\subsection{Ablation Study: Outlier Rejection Strategies}

We conduct an ablation study to evaluate the impact of different outlier rejection strategies on front-end SLAM accuracy. The four tested methods are: (1) \textbf{No-Rejection}: Accepting all correspondences without filtering. (2) \textbf{Fixed Threshold}: Using a constant global cutoff (e.g., 2.0 m). (3) \textbf{KISS-ICP Style}~\cite{vizzo2023kiss}: Adaptive frame-level threshold based on ego-motion and maximum LiDAR range. (4) \textbf{Ours (Point-Adaptive)}: Further refines thresholds per point, considering both ego-motion and individual point range to account for spatially varying motion distortion. These strategies were integrated into two representative LiDAR-inertial odometry frameworks: LIO-SAM~\cite{shan2020lio} and FAST-LIO~\cite{xu2022fast}.

\begin{table}[h]
    \centering
    {
    \caption{Ablation study comparing four outlier rejection strategies on front-end SLAM accuracy using FAST-LIO2 and LIO-SAM. Evaluated strategies include: (1) \textbf{No Rejection}, (2) \textbf{Fixed Threshold}, (3) \textbf{KISS-ICP Style}~\cite{vizzo2023kiss}, and (4) \textbf{Ours (Point-Adaptive)}. Results are reported across sequences from the GEODE and ECMD datasets.} 
    \label{tab:outlier_rejection_ablation}
    \resizebox{0.40\textwidth}{!}{
    \begin{tabular}{lcccccccc}
        \toprule
        \multirow{2}{*}{Algorithm} & \multicolumn{4}{c}{FAST-LIO2} \\
        \cmidrule(lr){2-5}
        & (1) & (2) & (3) & (4) \\
        \midrule
        \multicolumn{5}{c}{\textbf{GEODE}} \\
        \midrule
        Offroad2\_beta & 0.40 & 0.38 & 0.38 & \textbf{0.36} \\
        Offroad3\_beta & 0.26 & 0.23 & 0.27 & \textbf{0.19} \\
        Offroad6\_beta & 0.32 & 0.33 & 0.37 & \textbf{0.29} \\
        \midrule
        \toprule
        \multirow{2}{*}{Algorithm} & \multicolumn{4}{c}{LIO-SAM} \\
        \cmidrule(lr){2-5}
        & (1) & (2) & (3) & (4) \\
        \midrule
        \multicolumn{5}{c}{\textbf{ECMD}} \\
        \midrule
        Dense\_street\_day\_easy\_a & 0.71 & 0.70 & 0.71 & \textbf{0.69} \\
        Bridge\_day\_medium\_b & \textbf{1.12} & - & 1.17 & 1.14 \\
        Bridge\_day\_difficult\_a & 0.44 & 0.44 & 0.45 & \textbf{0.43} \\
        \bottomrule
    \end{tabular}
    }}
\end{table}

Table~\ref{tab:outlier_rejection_ablation} demonstrates that our point-adaptive strategy achieves superior accuracy across nearly all sequences and SLAM frameworks. By adapting thresholds per point, it effectively mitigates spatially variant distortions arising from ego-motion and range variation. The consistent performance gains in both filtering-based (FAST-LIO) and optimization-based (LIO-SAM) pipelines underscore the generalizability and efficiency of the proposed method.

\subsection{LiDAR-Inertial Odometry Performance}

Our LiDAR-IMU algorithm is integrated into a modified LIO-SAM framework and compared against several state-of-the-art methods, including FAST-LIO2~\cite{xu2022fast}, LIO-SAM~\cite{shan2020lio}, VoxelMap~\cite{yuan2022voxelmap}, and PV-LIO~\cite{jiang2022pvlio}. FAST-LIO2 adopts a tightly coupled IESKF, LIO-SAM is optimization-based, VoxelMap enhances localization through local geometric features, and PV-LIO incorporates covariance-aware weighting to improve estimation accuracy. We evaluate all methods on three diverse datasets to demonstrate the robustness and generalization capabilities of our approach in challenging environments.

\begin{table}[h]
    \small
    \centering
    \caption{Evaluation of LiDAR-Inertial SLAM using absolute pose error (APE, meters) on three benchmark datasets: GEODE, SubT-MRS, and ECMD.}
    \label{tab:lio_results_combined}
    \setlength{\tabcolsep}{2pt}
    \resizebox{0.48\textwidth}{!}{
    \begin{tabular}{lccccc}
        \toprule
        Sequence & FAST-LIO2 & LIO-SAM & PV-LIO & VoxelMap & D²-LIO (Ours) \\
        \midrule
        \multicolumn{6}{c}{\textbf{GEODE}} \\
        \midrule
        Tunnel-2    & 0.14 & 0.13 & 0.16 & 0.13 & \textbf{0.10} \\
        Tunnel-3    & 0.17 & 0.17 & 0.22 & 0.19 & \textbf{0.14} \\
        Offroad-1   & \textbf{0.12} & 0.15 & 0.26 & 0.19 & \textbf{0.12} \\
        Offroad-3   & 0.34 & -    & 0.24 & \textbf{0.22} & 0.75 \\
        Water-short & 0.29 & 0.32 & \textbf{0.12} & -    & 0.31 \\
        Flat Ground & -    & -    & -    & -    & -  \\
        \midrule
        \multicolumn{6}{c}{\textbf{SubT-MRS}} \\
        \midrule
        Final\_Challenge\_UGV1 & 1.74 & -    & -    & -    & \textbf{0.30} \\
        Urban\_Challenge\_UGV1 & 0.22 & 0.04 & 0.04 & -    & \textbf{0.03} \\
        Long\_Corridor\_1      & \textbf{1.08} & -    & -    & -    & 1.65 \\
        \midrule
        \multicolumn{6}{c}{\textbf{ECMD}} \\
        \midrule
        Dense\_street\_day\_easy\_a      & 0.45    & 0.71    & \textbf{0.34}    & 0.42    & 0.71  \\
        Highway\_day\_easy\_a      & 0.46    & -    & -    & 0.42    & \textbf{0.13}  \\
        Highway\_day\_medium\_a      & 0.50    & \textbf{0.29}    & -    & -    & \textbf{0.29}  \\
        Highway\_day\_medium\_b      & -    & -    & -    & \textbf{0.31}    & 0.35  \\
        Bridge\_day\_medium\_b  & -    & -    & -    & -    & \textbf{1.24}  \\
        Bridge\_day\_difficult\_a  & 1.35    & -    & -    & -    & \textbf{0.44}  \\
        Bridge\_day\_difficult\_b  & -    & -    & -    & -    & -  \\
        \bottomrule
    \end{tabular}
    }
\end{table}

In the GEODE dataset, we focus on two representative types of geometric degeneracy: long tunnels with repetitive structures causing self-similarity, and open offroad areas with sparse features. Our method effectively alleviates structural ambiguity, achieving the lowest errors in tunnel sequences while preserving competitive accuracy in feature-sparse settings—demonstrating robustness across diverse degradation modes. The SubT-MRS dataset introduces additional challenges such as mine-like tunnels, repetitive urban layouts, extended corridors, and multi-floor indoor environments, where limited LiDAR range and constrained sensor configurations intensify geometric degradation. In these settings, our approach maintains stable and accurate localization, particularly in long-duration sequences where other methods are prone to drift or failure. This robustness further extends to traffic-related degradation scenarios of ECMD, where our algorithm shows clear advantages—especially in moderately degraded highway and bridge environments—offering improved precision and resilience over existing solutions.

It is worth noting that the performance of our algorithm varies with the severity of geometric degradation. In mildly degraded environments—such as the GEODE tunnel and ECMD dense street—its advantages are comparatively marginal. However, under moderate degradation, exemplified by the ECMD highway and bridge, the method demonstrates notable robustness by effectively addressing localization challenges. Yet, in severely degenerated scenarios like the GEODE flat ground and parts of the ECMD bridge, LiDAR-IMU fusion alone proves inadequate for maintaining reliable localization, underscoring the fundamental limitations in the absence of richer environmental cues.

\subsection{ Eigenvalue Analysis and Degeneracy Compensation}

\begin{figure}[htp]
  \centering
  \includegraphics[width=0.45\textwidth]{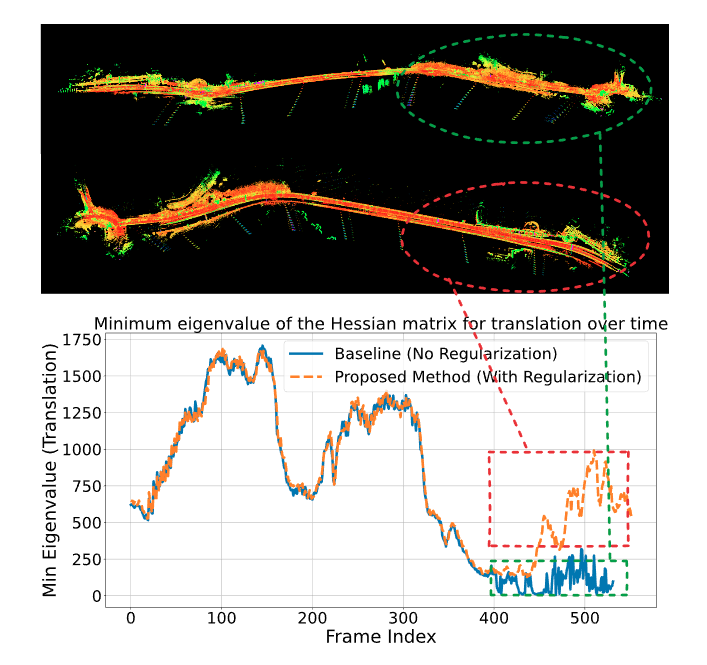}
  \captionsetup{justification=raggedright, singlelinecheck=false}
  \caption{Top: Point cloud trajectories of LIO with and without the proposed directional degeneracy-aware strategy. Bottom: Corresponding temporal evolution of the minimum eigenvalue of the translation Hessian, showing improved robustness with the proposed method.}
  \label{fig:PC_compare_with_Eigenvalue}
\end{figure}

To rigorously assess the impact of incorporating a directional degeneracy-aware residual term, we conducted controlled experiments on a highway scenario from the ECMD dataset. Specifically, we compare the performance of our system with and without the proposed pose estimation considering directional degeneracy strategy, as illustrated in Fig. \ref{fig:PC_compare_with_Eigenvalue}. The results reveal that, toward the end of the trajectory—where the scene exhibits insufficient geometric constraints—pose estimation without the additional residual term suffers from severe degradation, leading to significant localization drift. In stark contrast, the system augmented with the proposed residual maintains robust and accurate pose tracking, demonstrating its resilience under sudden degeneracy.

Furthermore, this performance difference is substantiated by the temporal evolution of the minimum eigenvalue of the translation-component Hessian. The comparison clearly indicates that the incorporation of the directional residual term adaptively elevates the smallest eigenvalues, thereby regularizing the optimization problem and enhancing its numerical stability. These findings collectively validate the effectiveness of our method in detecting and compensating for structural degeneracy, ultimately improving the robustness and reliability of LiDAR-IMU-based localization in challenging environments.

\section{CONCLUSIONS}

In this paper, we presented an enhanced LiDAR-IMU odometry framework that explicitly accounts for directional degradation during pose estimation. By incorporating a novel adaptive outlier rejection strategy and a degeneracy-aware optimization formulation, our method significantly improves robustness and accuracy in challenging environments. The integration of IMU preintegration covariance and scan-to-submap degeneracy analysis enables dynamic adjustment of optimization weighting, effectively constraining the solution in poorly conditioned directions. Both theoretical analysis and experimental results validate the effectiveness of the proposed approach in maintaining high-fidelity pose estimation under geometrically degenerate scenarios. Future work will focus on developing more robust and adaptive front-end matching strategies to further improve system resilience in complex and dynamic environments, particularly in cases of low-feature or repetitive geometries.

\addtolength{\textheight}{-12cm}   % This command serves to balance the column lengths

%%%%%%%%%%%%%%%%%%%%%%%%%%%%%%%%%%%%%%%%%%%%%%%%%%%%%%%%%%%%%%%%%%%%%%%%%%%%%%%%

\end{document}